\documentclass[journal]{IEEEtran}
\usepackage{graphicx}
\usepackage{amsmath}
\usepackage{subfigure}
\usepackage{url}
\usepackage{gensymb}
\usepackage{color}

\graphicspath{{figures/}}

\begin{document}

\title{Learning-Based Cost Functions for 3D and 4D Multi-Surface Multi-Object Segmentation of Knee MRI: Data from the Osteoarthritis Initiative}

\author{Satyananda Kashyap, Honghai Zhang, Karan Rao, and Milan Sonka,~\IEEEmembership{Fellow,~IEEE}

\thanks{
This research work was supported by NIH Grant No-R01 EB004640. The OAI is a public-private partnership comprised of five contracts (N01-AR-2-2258; N01-AR-2-2259; N01-AR-2-2260; N01-AR-2-2261; N01-AR-2-2262) funded by the National Institutes of Health, a branch of the Department of Health and Human Services, and conducted by the OAI Study Investigators. Private funding partners include Merck Research Laboratories; Novartis Pharmaceuticals Corporation, GlaxoSmithKline; and Pfizer, Inc. Private sector funding for the OAI is managed by the Foundation for the National Institutes of Health. This manuscript was prepared using an OAI public use data set and does not necessarily reflect the opinions or views of the OAI investigators, the NIH, or the private funding partners.

S.\ Kashyap, H.\ Zhang and M.\ Sonka are with the Dept.\ of Electrical and Computer Engineering, The University of Iowa, Iowa City IA 52242, USA 

K.\ Rao is with the University of Iowa Carver College of Medicine, Iowa City, IA 52242, USA}
}
\maketitle

\begin{abstract}
A fully automated knee MRI segmentation method to study osteoarthritis (OA) was developed using a novel hierarchical set of random forests (RF) classifiers to learn the appearance of cartilage regions and their boundaries. A neighborhood approximation forest is used first to provide contextual feature to the second-level RF classifier that also considers local features and produces location-specific costs for the layered optimal graph image segmentation of multiple objects and surfaces (LOGISMOS) framework. Double echo steady state (DESS) MRIs used in this work originated from the Osteoarthritis Initiative (OAI) study. Trained on 34 MRIs with varying degrees of OA, the performance of the learning-based method tested on 108 MRIs showed significant reduction in segmentation errors (\emph{p}$<$0.05) compared with the conventional gradient-based and single-stage RF-learned costs. The 3D LOGISMOS was extended to longitudinal-3D (4D) to simultaneously segment multiple follow-up visits of the same patient. As such, data from all time-points of the temporal sequence contribute information to a single optimal solution that utilizes both spatial 3D and temporal contexts. 4D LOGISMOS validation on 108 MRIs from baseline and 12 month follow-up scans of 54 patients showed significant reduction in segmentation errors (\emph{p}$<$0.01) compared to 3D. Finally, the potential of 4D LOGISMOS was further explored on the same 54 patients using 5 annual follow-up scans demonstrating a significant improvement of measuring cartilage thickness (\emph{p}$<$0.01) compared to the sequential 3D approach. 
\end{abstract}


\section{Introduction}
Knee osteoarthritis (OA) is a leading cause of disability in the United States, affecting over 10\% percent of the adult population, and accounting for over 600,000 knee replacements costing in excess of \$10 billion per year \cite{Zhang2010, Murphy2012, Cram2012, Kremers2015}. Although commonly characterized by joint pain and functional disability, OA is a heterogeneous disease with multiple risk factors and pathways contributing to variable disease symptoms and progression. The pathogenesis of OA is complex, with biomechanical joint stressors, inflammatory mediators, and structural changes to bone, ligaments, joint synovium, and menisci leading to articular cartilage degradation and loss \cite{Robinson2016, Lane2011}. No FDA-approved drugs are available which can forestall cartilage degeneration. Imaging biomarkers that are sensitive to structural changes are needed for early detection of OA in an effort to provide useful insights to clinical trials investigating disease modifying drugs. Magnetic resonance imaging (MRI) of the knee provides a complete 3D view of the knee anatomy unlike traditional 2D x-rays. Moreover, changes in cartilage morphology are detected earlier in MRIs than x-rays \cite{Sharma2008}.  In a longitudinal study by Raynauld et al.\ \cite{Raynauld2005}, MRI was used to reliably measure structural changes in arthritic knees and detect cartilage volume loss as early as 12 months from baseline. Balamoody et al. \cite{Balamoody2010} demonstrated that similar MRI protocols across different 3T scanners gave comparable morphologic results, further suggesting that MR imaging is a dependable tool for quantitative cartilage morphology analysis.

For analysis of the cartilage structures, accurate segmentation is a crucial first step. In clinical research there is a need for reproducible fully or highly automated segmentation as it offers consistent accuracy and speed over manual segmentation efforts. However, automated knee segmentation is challenging due to the thinning cartilage, appearance of osteophytes, bone marrow and cartilage lesions and surface fibrillation in MRI data. Several of these disease artifacts appear similar to the cartilage further increasing the segmentation complexity (Fig.~\ref{fig:4d}a).

Automated knee segmentation approaches previously reported in the literature were surveyed based on the employed methodology. Active shape and appearance models (ASM/AAM) were widely used in a variety of combinations such as deformable active shape models \cite{Fripp2010, Fripp2007}, volumetric and surface based appearance models \cite{Williams2010}, semi-automated initialization of active shape models \cite{Lynch2001}, and minimum descriptor length based group-wise image registration followed by active appearance model  segmentation \cite{Vincent2010}. In the presence of pathology, active shape/active appearance models need to be flexible enough to properly represent many local changes of shape/appearance, thus requiring large training-set sizes and yielding high-dimensional models. The associated model-training and computational complexity is already substantial in 3D and may become impractical if 4D longitudinal analysis is considered. Multi-atlas based methods were used in several algorithms such as \cite{Shan2014} with a non-local patch based fusion, Dam et al.\ \cite{Dam2015}  extended their multi-atlas based rigid registration by k-nearest neighbor voxel based classification in \cite{Folkesson2007}. Multi-atlas registration followed by locally weighted voting was reported in \cite{Lee2014},  multi-atlas based registration followed by outlier detection to segment the tibio-femoral joint in \cite{Tamez-Pena2012}. Since multi-atlas methods optimize a match between the overall template labeling as well as the individual labeled templates in the set, they may be sensitive to imperfect label templates in the presence of pathology. Both multi-atlas and ASM/AAM approaches may also suffer from converging to a locally rather than globally optimal solutions. Markov random fields were used to construct and optimize local image patches for the region and boundary probabilities from local shape and appearance information in \cite{Lee2011}. Graph cuts optimization was applied to the output of a hierarchical two-stage classifier, which was trained to identify the cartilage and  bone voxels \cite{Wang2014}. As the underlying multi-label graph cuts jointly consider independent classifier outputs for cartilage,  bone, and background, label-conflict-resolution  may be challenging in regions with multiple labels. In Zhang et al.\ \cite{Zhang2013}, four different MR acquisitions of the same patients were used to leverage the benefits of different sequence contrasts to model the spatial contextual information using support vector machines in a discrete random field framework. While a conceptually interesting approach, the need for multiple MR acquisitions limits the method's applicability.

Using cost functions designed as gradient-based features with heuristically determined weighting combination based on human expert knowledge (hand-tuned cost functions, Section II) limits the segmentation performance especially in the pathologic cases. Machine learning based methods can be used to identify the boundary properties of the anatomy of interest. Such learning-based cost function design has shown promise previously \cite{Brejl-TMI-00}. 

A major limitation of learning-based approaches is the time-consuming task of manually segmenting a large number of accurate examples used for training. Several post-processing interactive techniques were reported to ease the corrections, e.g., thin plate splines \cite{ross2009lung}, live wire techniques \cite{schenk2000efficient} and active shape models \cite{schwarz2008interactive} with the major issue being that they correct for the surfaces directly to match the object boundaries and are sometimes infeasible to be computed in close-to-real time.

We present a fully automated LOGISMOS segmentation algorithm that guarantees global optimality with respect to the cost functions provided. A novel hierarchical random forest (RF) classifiers was designed to provide the learned costs. An interactive correction method called just-enough interaction (JEI) \cite{Sun2013,Kashyap-IMIC2016} was used to substantially reduce the interaction time needed to prepare the training data in comparison with voxel-by-voxel editing approaches. In our method, two variations of the RF classifier were used --- a neighborhood approximation forests (NAF) followed by an RF classifier collected on a geometric graph. The novelty of this approach is that the NAF classifier gathers contextual and textural information from a global neighborhood of 3D image patches while the RF classifier collects local feature information along the geometric graph search columns. This combination enables training on contextual and textural features  both locally and globally thus improving segmentation accuracy.
 
OA as a disease progresses slowly in the early stages with the rate of cartilage loss accelerating with advancing disease \cite{Eckstein2014a}. Every patient in the OAI was enrolled in the longitudinal study with follow-up MRI scans done at regular clinical visits. The availability of multi-time-point information can be leveraged to provide additional information proving beneficial in patients with progressively worsening OA. Using the spatial and temporal contextual information from the neighboring time-points of the same patient helps reducing the inter-time-point variability ensuring that the cartilage losses are within physiologically feasible ranges. The LOGISMOS framework was extended to 4D to incorporate this information and simultaneously segment the multiple follow-up scans of the same patient. To the best of our knowledge, this is the first attempt at a 4D knee MRI segmentation algorithm incorporating the additional available information in a LOGISMOS framework.

The work reported here improves several aspects of the LOGISMOS algorithm for cartilage segmentation previously presented in \cite{Yin2010}. Our novel hierarchical random forest classifier outperforms the previous method, in which a single RF classifier did not account for the regionally-specific appearance of the surrounding menisci, muscle, bone and other anatomies. As a result, certain intensity profiles appearing in normal-cartilage regions of the knee also occurred in other local regions in a pathological case. This ambiguity resulted in the imperfect training of the classifier and contributed to segmentation inaccuracy.  Further, the LOGISMOS algorithm was extended to handle simultaneous multi-3D (4D) segmentation.

The remainder of the paper is organized as follows: Section \ref{LOGISMOS} introduces the LOGISMOS algorithm framework and the 3D graph construction. In Section \ref{cost_function} our proposed hierarchical RF classifier design is described in detail followed by the 4D LOGISMOS extension in Section \ref{4D}. The experimental design, validation, and discussion follow in Sections \ref{expMethods}, \ref{results} and \ref{conclusions}.

\subsection{LOGISMOS Segmentation}
\label{LOGISMOS}

LOGISMOS algorithm \cite{Yin2010} solves the simultaneous segmentation of multiple objects and surfaces in a graph-based framework with guarantee of global optimality. To ensure global optimality, the LOGISMOS algorithm has to satisfy two important criteria: a) The geometric graph must have non-intersecting column construction and b) the graph edge construction must satisfy the self-closure property (see \cite{Wu2005, Li2006PAMI} for proof). The focus of this paper is reporting a new method for cost function design and extending the 3D LOGISMOS to 4D. The workflow originally proposed in \cite{Yin2010} is maintained.

The algorithm is initialized by identifying the volume of interest (VOI) which uses an AdaBoost classifier \cite{Freund1997} trained on manually identified VOI's. The purpose of VOI detection was to localize smaller regions to reduce computation time. Further the VOI bounds are used for affine fitting of the mean femur and tibia bone shape mesh $S_0$. A patient specific bone mesh $S$ is computed by a single surface single object LOGISMOS segmentation using $S_0$. This step is necessary since the final segmentation of the bone and the cartilage surfaces is defined by this prior.

A geometric graph with non-intersecting columns representing each surface (bone and cartilage of femur and tibia) is built by mimicking the behavior of charged particles resulting in electric lines of force (ELF) based columns. The cost functions designed are assigned to the graph nodes of every column. The graph-based multiple surfaces and objects segmentation edge construction and optimization are described in \cite{Yin2010}. 

\section{Cost function design}
\label{cost_function}
LOGISMOS guarantees global optimality of its solution with respect to the cost functions provided. 

Simple hand-tuned cost functions were first designed to capture the bone to cartilage interface, which appeared as a strong dark to bright edge when traversing the image along the graph column from inside of the bone surface outwards to the bright cartilage. Directional 1D derivative operators gave costs based on the edges encountered along the search columns. For the cartilage, a weighted combination of the first and second derivative operators was used. This helps prevent interpreting cartilage inhomogeneities as edges. Although human expert designed cost functions are very effective at capturing the desired features there are limitations. 1) Choosing the correct weighting combination is challenging. 2) Same costs may not work for all parts of the anatomy. 3) The same anatomical objects (e.g., bone, cartilage) appear differently in pathological cases due to the loss of structure and/or the presence of lesions. 

RF classifiers \cite{Breiman2001} were used to provide better and location-specific cost functions. RF classifiers use the concept of bagging, where for each decision tree a random subset of features is chosen thereby reducing inter-decision-tree correlation,  which improves accuracy. A single RF classifier was used in \cite{Yin2010} where the limitation was that all the information used was localized along the graph columns. Because of this locality, features were unable to capture information of the neighborhood that appeared larger than a few nodes along the column. There are several anatomical features that appear locally like a cartilage boundary but when examined in a global neighborhood may be combined with pathology (e.g., synovial fluid, Fig.~\ref{fig:4d}). Further, the single RF classifier did not account for the regionally-specific appearance of the surrounding menisci, muscle, bone and other anatomies. As a  result, certain intensity profiles appearing in a specific normal-cartilage region of the knee also occurred in another local region in a pathological case. This discrepancy resulted in the improper training of the classifier and contributed to segmentation inaccuracy.

In this work, a combination of global and local contextual features was explored. Two RF based classifiers working in hierarchy were used to learn image appearance properties in the cartilage regions (Fig.\ \ref{fig:workflow_train}). The first stage used a NAF \footnote{Based on code available at: \url{http://www.nmr.mgh.harvard.edu/~enderk/software.html}} \cite{konukoglu2013} trained on example image patches. The output probability maps of the NAF were used with other image-based features for training each of the regionally specific second-stage RF classifiers. The second RF classifier collected features along the geometric graph columns. For the second RF classifier, the local regions of the knee were spatially clustered using $k$-means clustering to account for variable local anatomy appearances surrounding the cartilage (Fig.~\ref{fig:clusterRF}). Each of the spatially clustered regions was trained using a different RF classifier in the second stage. Disjoint training sets were used to help build a more realistic RF model based on actual NAF performance on unseen images.

\subsection{Neighborhood Approximation Forests}
NAF uses a random forest framework consisting of a several binary decision trees where each tree independently learns to predict the closest neighborhood. NAF uses the training set to learn which clusters of image patches have the most similar neighborhood structures based on a similarity criterion. The definition of similarity is application specific with the advantage of using a wide variety of distance-based metrics without modifying the core underlying approach. Upon training, NAF approximates the neighbors of an unseen test image based on this similarity criterion. 

For each tree in the forest, the training phase starts with the root node and continues to add new branches. Each node branch is trained to learn a set of binary tests which progressively partition the image into subsets with respect to the user-defined distance metric. The node split is optimized such that the user-defined distance function yields the most compact partitioning. 
For every tree in the forest, a subset of the entire feature set is chosen for training thereby improving generalization and producing an independent prediction. 

In this work, the NAF was trained on image patches where the pairwise distance function $\rho(I,J)$ captured the similarity of the image patches according to their segmentation labels. The distance function between training image patches were defined as $\rho(I,J)~=~\parallel seg(I) - seg(J) \parallel_{l_0}$ where $seg(.)$ is the segmentation label map for the corresponding image patch. The algorithm learned to group image patches that appeared similar to each other based on the segmentation similarity. 

\begin{figure}[htb]
	\centering
	{\includegraphics*[height=0.5\columnwidth,width=0.6\columnwidth]{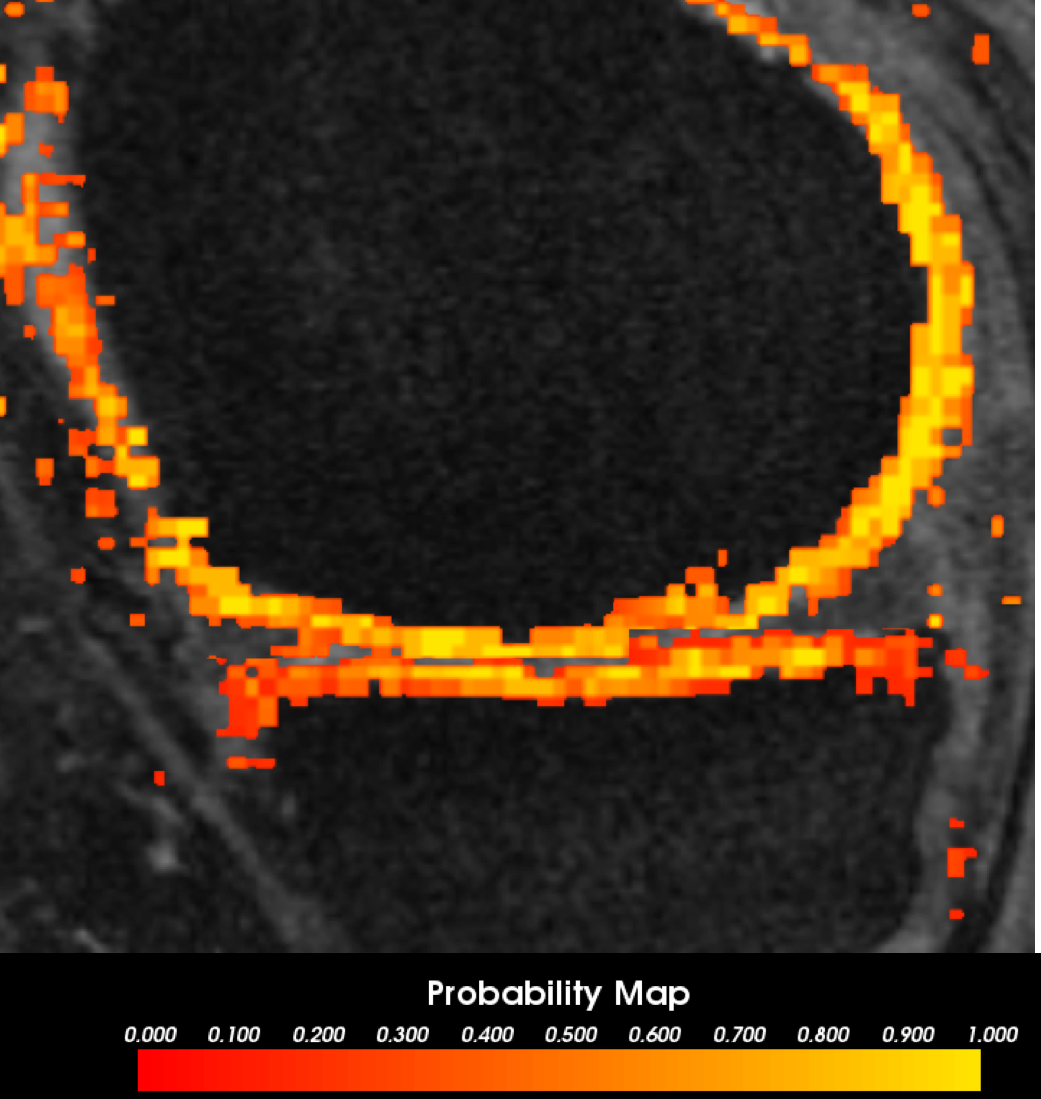}}
	\vspace{-.3cm}
	\caption{The output probability map of the NAF for an image overlaid on the image volume. The color map indicates the probability output values for each 3D voxel belonging to the femur (top) and tibia cartilage. The brighter yellow denotes higher probability while the red  indicates regions of lower probability of the voxel being a cartilage.}
	\label{fig:NAF}
		\vspace{-.4cm}
\end{figure}

The test image was partitioned into several smaller patches. Each patch passed through a trained NAF classifier. The output probability of all patches were combined to produce the final probability map (Fig.\ \ref{fig:NAF}). This was used as one of the inputs for the second clustered RF classifier. 

\subsection{Clustered Random Forest Classifier}
The second RF classifier was trained on features collected at every node along the search columns of the geometric graph. Bone mesh surfaces that were corrected using just-enough interaction (see Appendix \ref{jei}) were used for geometric graph construction during training. Positive example labels corresponded to the nearest cartilage mesh intersection along each graph column. The different features collected at each graph node are shown in Table \ref{featureLabel} with volumes of every feature computed individually. The value at each node point along the geometric ELF graph column was linearly interpolated from the closest point in the feature volume.

\begin{figure}[htb]
\vspace{-.2cm}
	\centering
	\includegraphics*[height=1in]{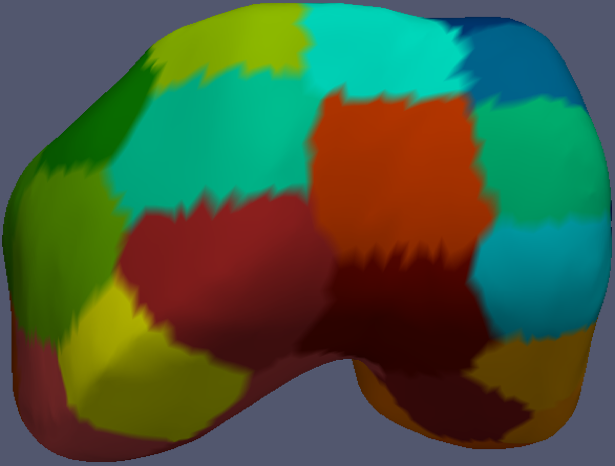} 
  \includegraphics*[height=1in]{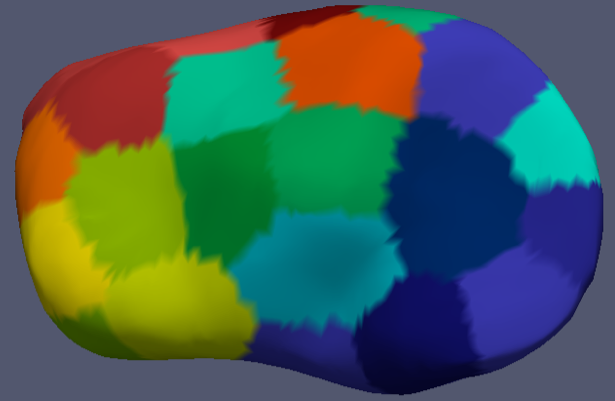}
  \vspace{-.2cm}
	\caption{Parcellation of the $S_0$ mesh using $k$-means clustering of the femur and tibia into 40 clusters each. Each colored region represents a different cluster that is trained using a separate RF classifier to account for the regionally-specific appearance of surrounding menisci, muscle bone and other anatomies.
}
	\label{fig:clusterRF}
	\vspace{-.4cm}
\end{figure}

To handle the variability of cartilage image intensities and the regionally specific surrounding anatomy, a $k$-means clustering algorithm was applied on the coordinates of femur and tibia mean shape $S_0$ mesh points. The spatial parcellation of the pre-segmented mesh in 40 clusters each was empirically designed (total 80, Fig.~\ref{fig:clusterRF}). Having too many clusters resulted in a reduced set of graph columns yielding insufficient sets of training examples. Having only a small number of clusters resulted in larger more inhomogeneous sets  while negatively affecting the quality of  classification. Training used a separate RF classifier for each of the clusters to learn variable cartilage appearance and consider surrounding anatomies. 

The probability response to the features along the search nodes in the testing datasets provided the node costs for LOGISMOS segmentation. 

\begin{table}[htb]
	\large
	\caption{List of features for the second RF classifier.}
	\label{featureLabel}
	\vspace{-.6cm}
	\begin{center}
	  \resizebox{\columnwidth}{!}{
		\begin{tabular}{l|l}
			\hline
			{\bf Index} & {\bf Description} \\ \hline
			1--9           & 3 eigenvalues of Hessian matrices on \\ & intensity image at $\sigma=0.5,1.0,2.0$ mm \cite{frangi1998multiscale}\\
			10--15         & Gaussian gradient on intensity and NAF \\&probability volumes at $\sigma=0.36,0.7,1.4$ mm \cite{zhao2009congenital}\\
			16--18         & Intensity, Gaussian smoothed intensity,\\ &and NAF probability volumes \\
			19--20         & Laplacian derivative of intensity volume\\ &at $\sigma=0.36,0.7$ mm \cite{zhao2009congenital} \\
			21             & Gabor texture feature \cite{daugman1985uncertainty} \\
			22--25         & Intensity statistics: mean, variance, skewness\\& and kurtosis of a 2 mm$^3$ region\\& centered around each graph node \cite{antonie2001application}\\
			26--28         & Haar features (1.5mm kernel) along \\& horizontal, vertical \& diagonal directions \cite{viola2004robust} \\
			29--30         & 1D directional gradient along the search column \\& direction on NAF probability and intensity volume\\
			\hline
		\end{tabular}
	  }
		\vspace{-.6cm}
  \end{center}
\end{table}

\section{4D Longitudinal Segmentation of Knee MRI}
\label{4D}
Along with the proposed learning-based cost functions, it is beneficial to leverage the contextual information from multiple time-points to improve the overall segmentation accuracy. Sequential yearly follow-up MRIs of the same knee provide spatial and temporal contextual information which helps to reduce the inter-time-point variability while ensuring physiologically feasible losses of cartilage during osteoarthritis progression. The crucial first step in 4D segmentation is to register the pre-segmented mesh surfaces and the respective images across time-points to establish correspondences temporally between the similar regions of the knee in 4D. A rigid registration was used given that only slight changes in bone shape occur with disease progression. The same configuration is maintained across all time-points in terms of the geometric graph parameters and topology. Enforcing contextual information temporally is done by linking adjacent time-points using  inter-time-point edges in the underlying 4D LOGISMOS graph.

\subsection{Establishing Temporal Correspondences}

\begin{figure}[htb]
	\includegraphics*[width=0.49\columnwidth]{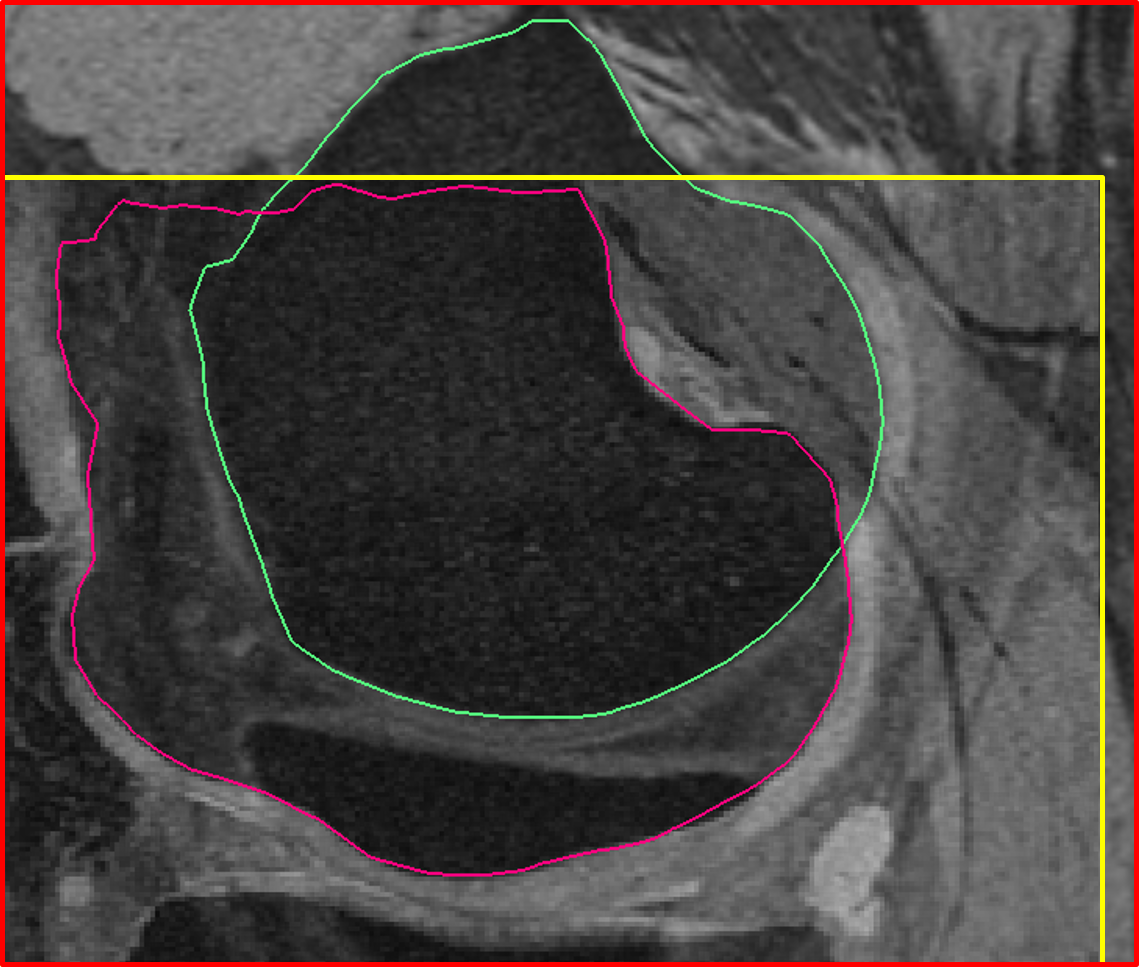} 
	\includegraphics*[width=0.49\columnwidth,height=3.7cm]{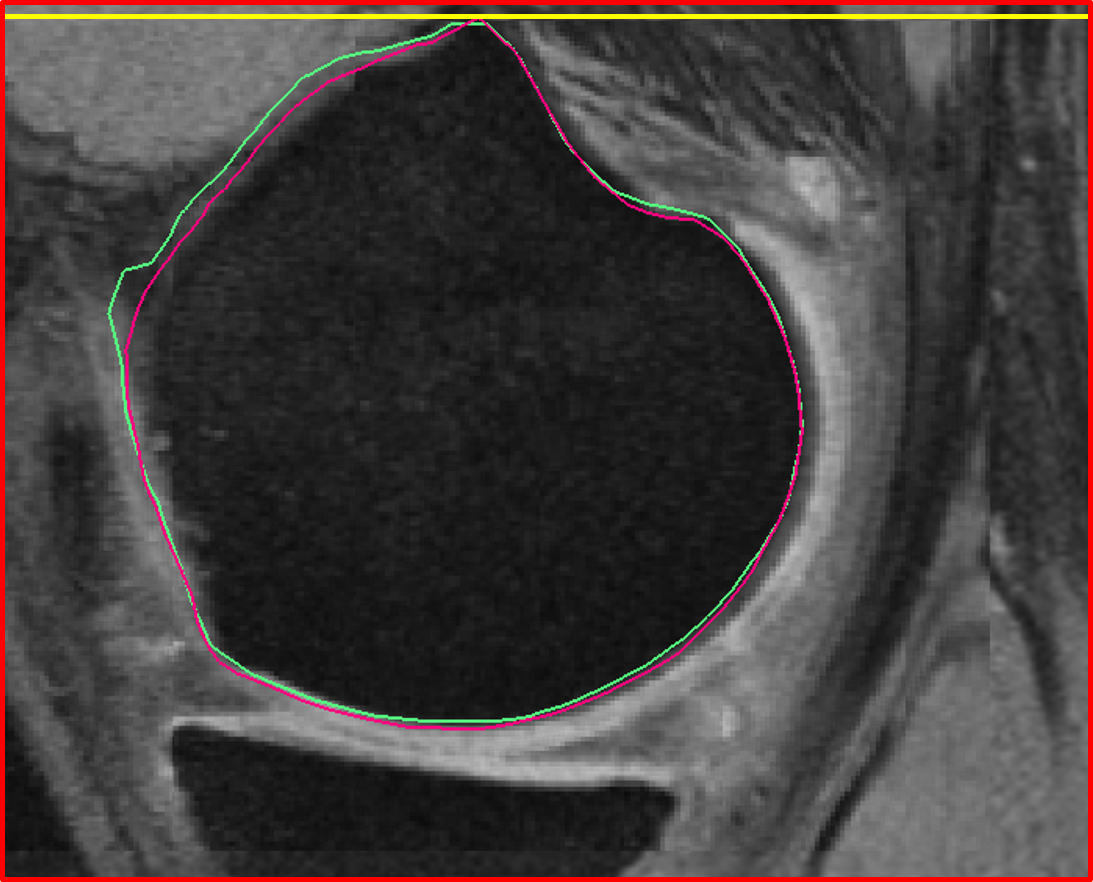}
	\vspace{-.3cm}
	\caption{Establishing point correspondences between time-points after pre-segmentation. The meshes are registered using ICP to establish column correspondences. The same transformation is applied to the corresponding volumes. (a)~Two time-points before registration. The meshes are colored differently and the volume border is highlighted in yellow to indicate its positioning. (b)~Meshes registered after ICP.}
	\vspace{-.3cm}
	\label{fig:icp}
\end{figure}

After pre-segmentation of each of the time-points, iterative closest points (ICP) algorithm was used to register the pre-segmented mesh surfaces. With a large translational or rotational movement between the two-time-points, there is a tendency to mismatch the surfaces, i.e., femur matched to the tibia or vice versa if they were the closest in terms of the least squares optimization. To prevent the mismatch, a two step registration was employed. The first step used the femur mesh only for ICP registration. This transform matrix was applied on both the femur and tibia meshes. After the first transform ensured that femoral and tibial surfaces were reasonably aligned, the ICP registration was run again using both the femur and tibia points together as a unified point cloud to further refine the registration. This ensured that vertex to vertex correspondences and thereby column correspondences were established. Fig.\ \ref{fig:icp} illustrates the establishment of correspondence between two time-points. The same two-step transformation matrices were applied to the entire images. 

\subsection{Incorporating Inter-Time Point Context}
For the 4D segmentation, the datasets at each time-point have their respective 3D multi-surface multi-object constraints enforced as described earlier in \cite{Yin2010}. In order to enforce the contextual constraints across adjacent time-points, inter-time-point edges are constructed connecting nodes in every column with the corresponding columns on the registered surfaces. Using the prior information we can set the different minimum and maximum permissible limits for the bones and cartilages respectively so that cartilage thickness changes are within a physiologically feasible range. The edges introduced define inter-time-point min and max ($\delta_{tmin}, \delta_{tmax}$) where the inter-time-point maximum allowed change ($\delta_{tmax}$) was set to 0.6 mm/year based on available clinical literature \cite{Eckstein2014a}. The minimum $\delta_{tmin}$ was set to zero.

The inter-time-point edges $E^{t}$ between neighboring time-points $t_1$, $t_2$ are constructed for all nodes of the graph:
\begin{align*}
E^{t} = \{\langle v_i(t_1,n_{cart},k), v_i(t_2,n_{cart},k-\delta_{tmin})\rangle\}\\
\cup \{\langle v_i(t_1,n_{cart},k-\delta_{tmax}), v_i(t_2,n_{cart},k)\rangle\}\\
\cup \{\langle v_i(t_1,n_{bone},k), v_i(t_2,n_{bone},k - \delta_{tmin})\rangle\}\\
\cup  \{\langle v_i(t_1,n_{bone},k-\delta_{tmax}), v_i(t_2,n_{bone},k)\rangle\},
\end{align*}
where $v_i(t,n,k)$ represents the $k$th node on the $i$th column, the inter-time-point edges are constructed between nodes $v_i(t_1,n,k), v_i(t_2,n,k)$ of longitudinally corresponding columns $t_1$, $t_2$ for every $n \in \{cart,bone\}$. After the graph edges and nodes have been constructed, the 4D LOGISMOS segmentation problem is solved by computing a minimum $\emph{s-t}$ cut in a derived edge weighted digraph \cite{Boykov2004}. 

\section{Experimental Methods} 
\label{expMethods}
Knee MRI data used for evaluation originated from the Osteoarthritis Initiative (OAI) database available for public access\footnote{\url{https://oai.epi-ucsf.org/datarelease/}}. All MR acquisitions used double echo steady state (DESS) pulse sequence with in-plane resolution of 0.36$\times$0.36mm and a slice spacing of 0.7mm, resulting in image volumes with 384 $\times$ 384 $\times$ 160 voxels. 88 datasets at baseline (BL) and 12-months follow-up (12M) scans (176 3D MRIs in total) were used for which manually defined independent standards are available via OAI. BL images of 34 patients were used for training the hierarchical RF classifiers. The 12M data of the same patients were excluded from the testing set given the similarity of appearance resulting in a testing set of 54 MRIs at BL and 12M respectively (total of 108 MRIs).

\begin{figure}[htb]
	\centering
	\includegraphics[width=\columnwidth]{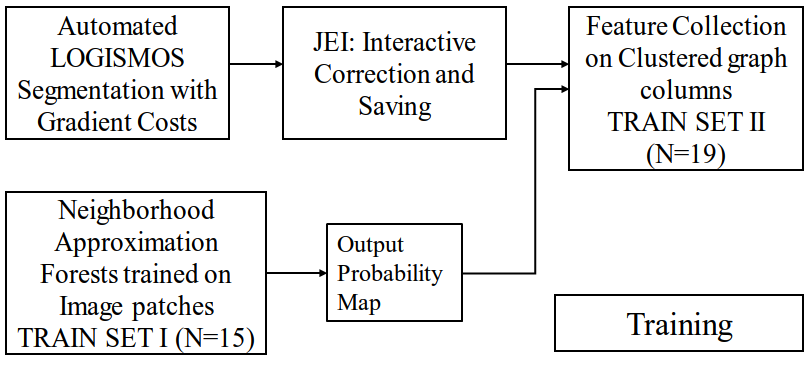}
	\vspace{-.3cm}
	\caption{Learning-based workflow for training the hierarchical RF classifiers.}
	\label{fig:workflow_train}
\end{figure}

Fig.\ \ref{fig:workflow_train} shows the training workflow for the hierarchical RF classifiers. For bone surface segmentation, the initially employed gradient-based costs were very robust and remained unchanged. For learning the cartilage-surface costs, the 34 patients were divided into two training sets with 15 and 19 patients each that were used to train the NAF and the clustered RF classifier respectively. The OAI provided independent standards had the cartilage and bone boundaries only in the region of articular cartilage (Fig.\ \ref{fig:qual}a). However the LOGISMOS segmentation and further analysis required the entire bone structure. The datasets used for training the clustered RF classifier were first inspected and JEI edited (see Appendix \ref{jei} for details). 

\begin{figure}[htb]
	\centering
	\vspace{-.3cm}
	\includegraphics[width=0.8\columnwidth]{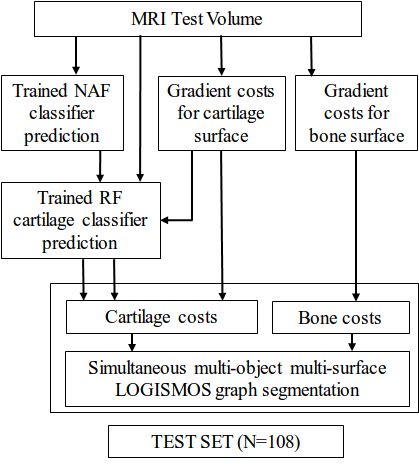}
  \vspace{-.3cm}
	\caption{Testing workflow to compare the clustered random forest classifiers with the existing methods on a set of 108 patients.}
	\vspace{-.3cm}
	\label{fig:workflow_test}
\end{figure}

\begin{table}[htb]
	\caption{Parameters used for graph construction.}
	\centering
	\label{graph-parameters}
	\begin{tabular}{lccccc}
		\hline
		& Inter-surface & Inter-object & Smoothness & Column size \\
		& max (mm)   & max (mm)  & (mm)    & (mm)     \\
		\hline
		Learned cost  & 6            & 18          & 0.6          & 18.15 \\
		Gradient cost & 4            & 12           & 0.4         & 12.2   \\
		\hline
		& & & & \\
		\multicolumn{5}{l}{*Minimum inter-surface, inter-object and inter-time-point separations are zero.}
	\end{tabular}
\end{table}

All image volumes were first LOGISMOS-segmented using gradient costs. The geometric graphs had 8006 and 8002 graph columns for the femur and tibia objects, respectively. The graph parameters are listed in Table \ref{graph-parameters} with the additional inter-time-point constraint set at 0.6 mm for the 4D LOGISMOS. 

The NAF features were sampled over 15 datasets with 1521 samples per image patch. Due to the imbalance in the negative to positive labels ratio, a neighborhood around the cartilage was marked as negative examples. A set of 200 trees with 40,000 image patches were used as inputs to the training. The second RF classifier was trained on 19 JEI corrected datasets with every node along the ELF search column collecting 30 features (Table~\ref{featureLabel}). A total of 80 RF classifiers (one per cluster) were trained with 800 trees per each forest. 

OA is characterized by thickness losses in subregions of the surface. Analyzing the entire cartilage structure may not be sensitive to these localized and regionally smaller changes. Therefore a consensus based nomenclature was proposed by clinicians identifying regions of the cartilage most likely affected by OA \cite{Eckstein2006a}. The existing methodologies in the literature require manual initialization for the sub-plate analysis \cite{Wirth2008,Wirth2010}. We developed a fully automated nomenclature-compliant sub-plate analysis algorithm. The algorithm subdivided the segmented surface by automated identification of the trochlear notch anatomy and implicit cutting plane geometry (Appendix \ref{sub-plate}). All of our experimental results were subplate analyzed and reported for the entire femur and tibia cartilage regions, central load-bearing lateral femur/tibia (cLF/cLT), and central load-bearing medial femur/tibia (cMF/cMT).

\begin{figure}[htb]
	\centering
	\includegraphics[width=0.9\columnwidth]{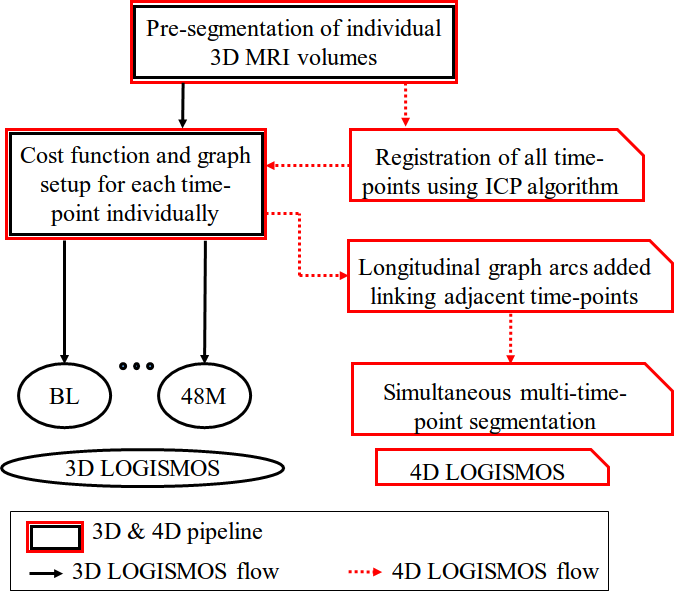}
	\caption{Comparison of 3D and 4D LOGISMOS pipelines. The black and red arrows highlight the flow of 3D and 4D LOGISMOS algorithms respectively.}
	\label{fig:3Dvs4D-flowchart}
\end{figure}

The experimental results are reported as signed and unsigned errors. The signed errors indicate the measurement bias with the positive errors denoting underestimation and the negative errors denoting overestimation with respect to the independent standard. The unsigned errors indicate the variability of the measurement. The error measurements were collected in regions for which the independent standard was available. The first experiment was to validate the accuracy of the different cost functions in 3D LOGISMOS segmentation. The costs by the proposed hierarchical RF classifier were compared against the gradient costs. To evaluate the benefits of adding the NAF stage, a single RF classifier was trained on the same 19 patient datasets excluding the NAF features while maintaining the same graph parameters as the hierarchical classifier system. The testing workflow is shown in Fig.~\ref{fig:workflow_test}.
The second experiment simultaneously 4D segmented the 54 knees using the proposed hierarchical costs and compared against the 108 individually 3D LOGISMOS segmented results using the same segmentation cost functions. It is designed to reveal the benefits of adding contextual information between the adjacent time-points. Given that OA is a slowly progressing disease, the expectation was that there would not be a significant appearance of artifacts or thickness losses over a one year study period. Fig.\ \ref{fig:3Dvs4D-flowchart} shows the flowchart highlighting the key differences between the 3D and 4D LOGISMOS pipelines. 

To demonstrate the benefits of 4D analysis over a longer study period, the same 54 patient datasets with BL, 12M, 24M, 36M and 48M follow-up scans were used resulting in a total of 237 3D MRIs analyzed for the third experiment (out of 270 due to subjects missing some follow-up visits). Since OAI-provided  independent standards were available only for BL and 12 M, a new independent standard was defined in house by an expert analyst using JEI approach (Appendix A).  To highlight the improvements, we compared the maximum cartilage thickness errors for 3D and 4D across all time-points and determined statistical significance of their differences.

\section{Results}
\label{results}
Table \ref{bone_error} shows surface positioning errors for the bone segmentation with the signed errors being sub-voxel accurate. Since high bone-segmentation accuracy was already achieved using the gradient-based costs, a learning based cost function design was not necessary. Although we saw presence of bone lesions with progression of OA, the gradient-based cost was sufficiently robust. Thus all the different methodologies in both 3D and 4D LOGISMOS used the same gradient-based cost functions for the bone.

\begin{table}[htb]
	\centering
	\caption{Bone segmentation errors (in mm) using the gradient-based cost functions.}
	\label{bone_error}
	\begin{tabular}{l|cc}
		\hline
		& Signed & Unsigned \\
		\hline
		Femur & -0.19$\pm$0.07 & 0.40$\pm$0.06 \\
		Tibia & -0.13$\pm$0.09 &  0.39$\pm$0.09 \\
		\hline
	\end{tabular}
\end{table}

Catilage surface positioning errors (compared against independent standard) using the hierarchical classifier, gradient cost, and single-stage RF classifier are reported. Each resulting surface from the above methods were sub-plate analyzed. Table \ref{error1} shows a significant reduction in signed and unsigned errors for all the femoral sub-plates ($p\ll0.001$). Significant reduction in unsigned errors over the segmentation using gradient costs was seen for all the tibial sub-plates while the signed errors showed a significant reduction of errors over the gradient-based costs on all sub-plates except the central medial tibia (cMT). Comparison with the single-stage RF classifier saw significant reduction of unsigned errors in the whole tibia and central medial tibia (cMT) plates. 

\begin{figure}[htb]
  \centering
	\begin{tabular}{cc}
		\multicolumn{2}{c}{\includegraphics[width=0.4\columnwidth]{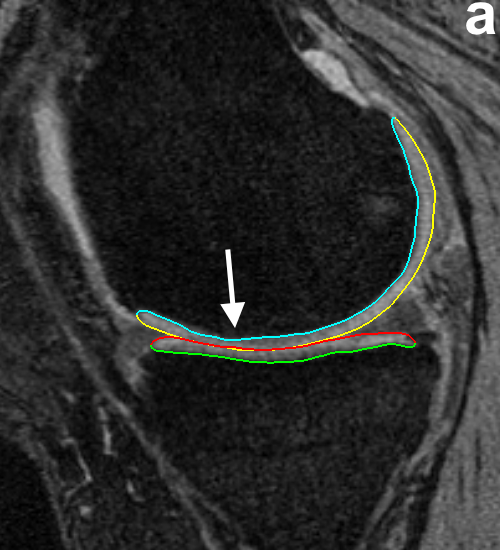}}\\
		\includegraphics[width=0.4\columnwidth]{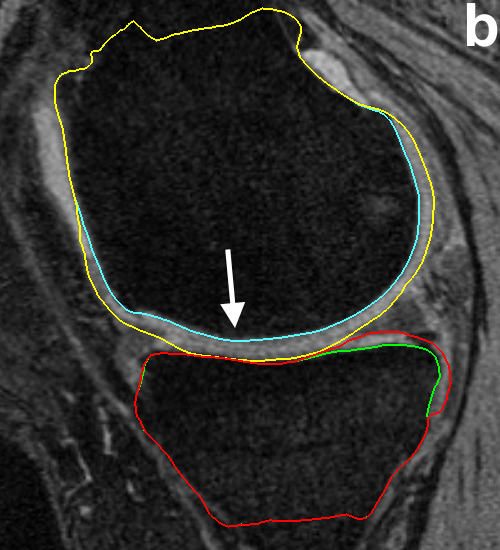} &
		\includegraphics[width=0.4\columnwidth]{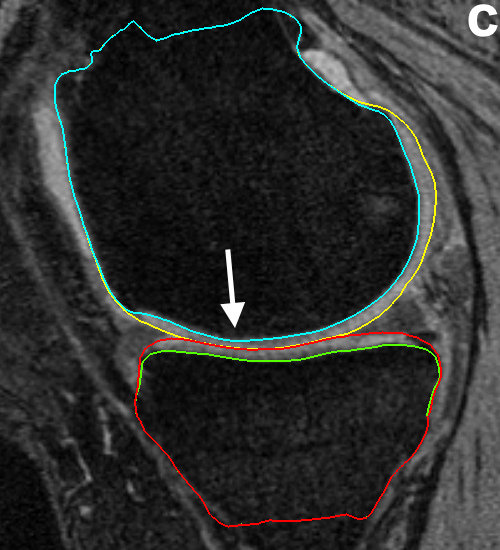} \\
	\end{tabular}
	\caption{Segmentation accuracy in a representative subject.
	(a)~Independent standard.
	(b)~Gradient-costs LOGISMOS segmentation.
	(c)~Learned-costs LOGISMOS segmentation. Region marked by the arrow shows clear improvement in the segmentation quality when using the learned costs.}
	\label{fig:qual}
\end{figure}

Fig.\ \ref{fig:qual} qualitatively compares the segmentation accuracies between the gradient-based costs and the hierarchical classifier with respect to the independent standard. Both the femur and tibia are shown with their respective bone and cartilage segmentations showing good agreement between learning-based segmentation and the independent standard.  

\begin{table*}[htb]
	\centering
	\caption{Cartilage surface positioning errors (in mm) achieved by hierarchical classifier, gradient cost and single-stage RF classifier using 3D LOGISMOS algorithm. Paired $t$-test comparisons are reported for (NAF+RF vs. Gradient) and (NAF+RF vs. RF only)}
	\label{error1}
	\begin{tabular}{lr|rr|rr}
		\hline
		\bf n=108 & NAF+RF           & Gradient          & $p$-value & RF only & $p$-value       \\ 
		\hline                                           
		Femur signed   & \bf -0.01$\pm$0.18 &-0.31$\pm$0.24     & $\ll0.001$  & -0.10$\pm$0.17& $\ll0.001$  \\ 
		Femur unsigned & \bf 0.55$\pm$0.11 & 0.69$\pm$0.13     & $\ll0.001$  &  0.56$\pm$0.10& $\ll0.001$  \\ 
		\hline     
		cMF signed   & \bf -0.04$\pm$0.29 &-0.38$\pm$0.58     & $\ll0.001$  & -0.11$\pm$0.27& $\ll0.001$  \\ 
		cMF unsigned & \bf 0.52$\pm$0.16 & 0.78$\pm$0.35     & $\ll0.001$  &  0.55$\pm$0.17& $\ll0.001$  \\ 
		\hline
		cLF signed   & \bf -0.26$\pm$0.24 &-0.52$\pm$0.35     & $\ll0.001$  & -0.36$\pm$0.20& $\ll0.001$  \\ 
		cLF unsigned & \bf 0.42$\pm$0.12 & 0.65$\pm$0.20     & $\ll0.001$  &  0.47$\pm$0.11& $\ll0.001$  \\ 
		\hline
		Tibia signed   &    \bf 0.06$\pm$0.17 & -0.11$\pm$0.35 & $\ll0.001$      & 0.11$\pm$0.22& $\ll0.001$  \\
		Tibia unsigned & \bf 0.60$\pm$0.14 &     0.79$\pm$0.20 & $\ll0.001$  & 0.62$\pm$0.18& $\ll0.001$  \\
		\hline
		cMT signed   & -0.15$\pm$0.31 &-0.25$\pm$0.77     & $0.193$  & \bf -0.09$\pm$0.34& $0.003$ \\ 
		cMT unsigned & \bf 0.52$\pm$0.20 & 0.92$\pm$0.41     & $\ll0.001$  &  0.58$\pm$0.22& $\ll0.001$ \\ 
		\hline
		cLT signed   & \bf -0.03$\pm$0.32 &0.36$\pm$1.12     & $\ll0.001$  & -0.01$\pm$0.32& $0.16$ \\ 
		cLT unsigned & \bf 0.46$\pm$0.17 & 0.79$\pm$0.99     & $\ll0.001$  &  0.47$\pm$0.18& $0.31$ \\ 
		\hline 
		\multicolumn{6}{l}{*Bold indicates statistically significant reduction in positioning errors.}
	\end{tabular}
\end{table*}

Table \ref{error-4Dvs3D} compares signed and unsigned border positioning errors with respect to the independent standard for 3D and 4D LOGISMOS segmentation at BL and 12M follow-up respectively. As expected, the overall segmentation accuracy is almost the same between the two methods except for a few of the plate regions highlighted in bold showing statistically significant differences over 3D ($p<0.05$).

Table VI shows the performance improvement obtained by the  4D LOGISMOS over 3D for a longer study period of five years (BL–--48 M, 5 time-points). To demonstrate improvements in local accuracy, Table VI gives the unsigned  errors of cartilage thickness averaged for three quantiles within 90-100\% of the largest errors  per patient combined for all available time-points for each sub-plate. These unsigned maximum thickness errors  were statistically significantly  smaller in each of cMF, cMT and cLT  for 4D LOGISMOS compared with separate 3D analyses.

\begin{figure}[htb]
  \centering
	\begin{tabular}{cc}
		\multicolumn{2}{c}{\includegraphics[width=0.4\columnwidth]{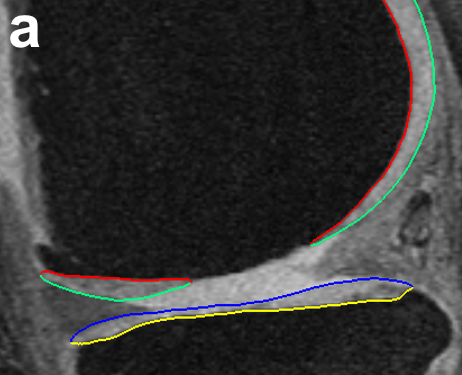}}\\
		\includegraphics[width=0.4\columnwidth]{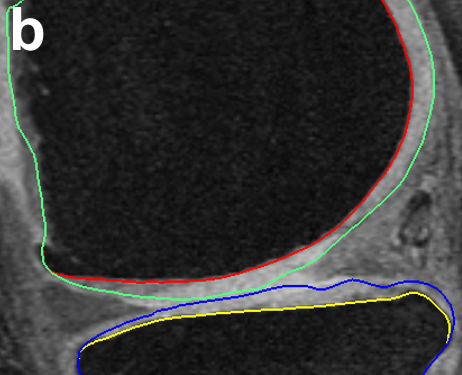} &
		\includegraphics[width=0.4\columnwidth]{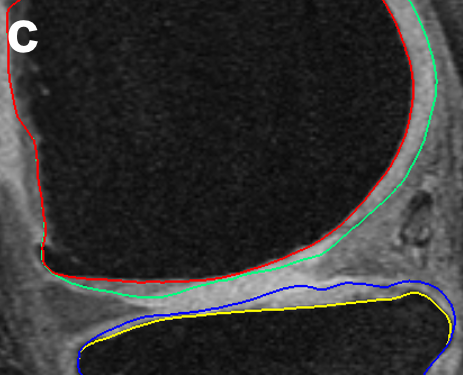} \\
	\end{tabular}
	\caption{Qualitative improvement of 4D segmentation versus 3D for a patient with severe osteoarthritis. Parts of the region between the femur and tibia exhibit synovial fluid leakage, the appearance and texture of which mimic that of a cartilage. (a) Independent standard, (b) 3D segmentation, (c) 4D segmentation. }
	\label{fig:4d}
\end{figure}

Fig.\ \ref{fig:4d} qualitatively shows the improvement of 4D LOGISMOS over 3D. Note the lack of an obvious edge distinguishing the tibia and femur cartilage. Using the spatial and temporal contextual information from all the time-points,  the 4D method is able to correctly position the cartilage for the tibia and femur despite the lack of image information locally. 

\begin{table*}[htb]
  \caption{Cartilage surface positioning errors (in mm) of 4D versus 3D LOGISMOS.}
  \label{error-4Dvs3D}
  \centering
  \begin{tabular}{l|rrr|rrr}
  \hline
                  & \multicolumn{3}{c|}{\bf n=54 Baseline} & \multicolumn{3}{c}{\bf n=54 12 months} \\
                  & \multicolumn{1}{c}{4D} & \multicolumn{1}{c}{3D} & $p$-value & \multicolumn{1}{c}{4D} & \multicolumn{1}{c}{3D} & $p$-value \\ \hline
  Femur signed    & \bf  0.01$\pm$0.18  & 0.01$\pm$0.19     & 0.027      & -0.02$\pm$0.17    &-0.04$\pm$0.17 & 0.062 \\ 
  Femur unsigned  & \bf  0.53$\pm$0.11  & 0.54$\pm$0.11     & $\ll$0.001 &  0.55$\pm$0.11    & 0.55$\pm$0.11 & 0.601 \\ \hline
  cMF signed      & \bf -0.01$\pm$0.25  &-0.03$\pm$0.27     & 0.015      & \bf 0.02$\pm$0.27 &-0.04$\pm$0.30 & 0.016 \\
  cMF unsigned    &  0.51$\pm$0.16      & 0.52$\pm$0.17     & 0.092      &  0.52$\pm$0.15    & 0.53$\pm$0.16 & 0.614 \\ \hline
  cLF signed      & -0.26$\pm$0.22      &-0.25$\pm$0.23     & 0.108      & -0.30$\pm$0.21    &-0.30$\pm$0.23 & 0.853 \\
  cLF unsigned    &  0.42$\pm$0.10      & 0.42$\pm$0.10     & 0.790      &  0.43$\pm$0.12    & 0.44$\pm$0.13 & 0.318 \\ \hline
  Tibia signed    &  0.08$\pm$0.17      &\bf 0.07$\pm$0.16  & 0.039      &  0.07$\pm$0.19    & 0.06$\pm$0.18 & 0.464 \\
  Tibia unsigned  &  0.59$\pm$0.14      & 0.60$\pm$0.14     & 0.140      &  0.60$\pm$0.16    & 0.60$\pm$0.14 & 0.310 \\ \hline
  cMT signed      & -0.13$\pm$0.28      &-0.14$\pm$0.29     & 0.125      & -0.14$\pm$0.29    &-0.15$\pm$0.31 & 0.619 \\
  cMT unsigned    & \bf 0.50$\pm$0.18   & 0.51$\pm$0.20     & 0.039      &  0.52$\pm$0.17    & 0.53$\pm$0.17 & 0.341 \\ \hline
  cLT signed      &  0.00$\pm$0.30      & 0.00$\pm$0.31     & 0.338      & -0.06$\pm$0.31    &-0.06$\pm$0.30 & 0.629 \\
  cLT unsigned    & \bf 0.45$\pm$0.17   & 0.46$\pm$0.18     & $\ll$0.001 &  0.45$\pm$0.16    & 0.44$\pm$0.16 & 0.256 \\ 
  \hline
	\multicolumn{7}{l}{*Bold indicates statistically significant reduction in positioning errors.}
  \end{tabular}
\end{table*}

\begin{table*}[htb]
	\caption{Longitudinal max thickness error values of 4D versus 3D LOGISMOS.}
	\label{max_thicknessError-4Dvs3D}
	\centering
	\begin{tabular}{l|rrr|rrr|rrr}
		\hline
    & \multicolumn{3}{c|}{\bf 2\% Max Average Errors (98-100\%)} & \multicolumn{3}{c|}{\bf 5\% Max Average Errors (95-100\%)} & \multicolumn{3}{c}{\bf 10\% Max Average Errors (90-100\%)} \\
		& \multicolumn{1}{c}{4D} & \multicolumn{1}{c}{3D} & $p$-value & \multicolumn{1}{c}{4D} & \multicolumn{1}{c}{3D} & $p$-value & \multicolumn{1}{c}{4D} & \multicolumn{1}{c}{3D} & $p$-value \\ \hline
    cMF unsigned    &  \bf 0.76$\pm$0.60 & 1.11$\pm$0.63     & $\ll$0.001 & \bf 0.57$\pm$0.43    & 0.87$\pm$0.41 & $\ll$0.001  &  \bf 0.43$\pm$0.32    & 0.68$\pm$0.29 & $\ll$0.001 \\ \hline
    cLF unsigned    &  1.46$\pm$2.36     & 1.67$\pm$1.82     & 0.610      &  1.07$\pm$1.73    & 1.16$\pm$1.02    & 0.74        &  0.74$\pm$0.95    & 0.84$\pm$0.55 & 0.528 \\ \hline
    cMT unsigned    & \bf 1.28$\pm$0.47  & 1.78$\pm$1.36     & 0.008      &  \bf 1.09$\pm$0.42    & 1.48$\pm$1.00 & 0.004      &  \bf 0.92$\pm$0.38    & 1.19$\pm$0.63 & $\ll$0.001 \\ \hline
    cLT unsigned    & \bf 1.29$\pm$0.47  & 1.40$\pm$0.49     & $\ll$0.001 &  \bf 1.09$\pm$0.40    & 1.17$\pm$0.41 & $\ll$0.001 &  \bf 0.90$\pm$0.33    & 0.98$\pm$0.34 & $\ll$0.001 \\ 
    \hline
		\multicolumn{7}{l}{*Bold indicates statistically significant reduction in positioning errors.}
	\end{tabular}
\end{table*}

\section{Discussion and Conclusion}
\label{conclusions}

A novel method for learning segmentation cost functions using  hierarchical RF classifiers  was reported and its performance was demonstrated in 3D and 4D LOGISMOS segmentation of knee MR images. The preparation of the training data was done using our just-enough interaction (JEI) approach, which provided fast and accurate post-processing correction. To minimize the impact of the underlying automated LOGISMOS segmentation on the independent standard used for training and evaluation, a great level of flexibility was embedded in the graph construction so that the expert analyst could freely modify the surfaces as needed and thus create unbiased corrections of the knee surfaces.

Using the single RF classifier improved the segmentation accuracy over the initial gradient-based costs. In addition, using a hierarchy of NAF and RF stages showed further statistically significant improvement in segmentation performance. The previous LOGISMOS-based method \cite{Yin2010} used a single-stage RF with the graph constructed in the image-voxel space and a subset of current features contributing to the cost function (Table~\ref{featureLabel}) and it placed third on the publicly available SKI10 challenge \cite{heimann2010segmentation}. The method reported here shows a significant  improvement. Further comparison with the various other methods would inevitably be imperfect due to the different testing datasets utilized. In the OAI cohort, our method achieved high accuracy in comparison with the publicly available independent standard. Training of the hierarchical RF classifier and the subsequent 3D/4D LOGISMOS analyses ran on a Linux system with CentOS-7.3, 128 GB RAM. Once trained, the average segmentation times were 34 and 22~minutes per time-point for individual 3D and simultaneous 4D LOGISMOS, respectively.

Leveraging the benefits of the available multi-time-point MRI datasets for the OA patients, we designed a 4D extension of the LOGISMOS segmentation framework. The added temporal inter-time-point contextual constraints helped constrain the changes of cartilage thickness to  what was physiologically feasible. This benefit of using information from the previous time-points along with the new cost function design helped notably improve the segmentation accuracy. Further, the advantages of the 4D LOGISMOS were demonstrated on an OAI longitudinal patient dataset with 5 imaging sessions from BL to 48 month. Our sub-plate detection algorithm was used to study the accuracy of segmentation on the load-bearing regions of the knee and provided good performance insight. The robustness of the automated sub-plate division was demonstrated showing consistent measurements between different surfaces used to initialize the automated extraction of the sub-plate load-bearing regions. 

The combination of LOGISMOS and learning-based methods has been  very effective. Other learning-based techniques such as deep learning may further improve the performance once more ground-truth data become available to determine further benefits of such approaches. 4D LOGISMOS also offers a platform for future clinically-oriented work, such as investigating the rate of cartilage degradation and its impact on OA progression, and identifying cartilage thickness that correlates to disease severity and/or predicts the need for joint replacement.
Assessing reproducibility of the reported method is left for future work and the OAI scan-rescan dataset is well-suited for this task. Patellar segmentation was not considered in the work reported here due to the limited validation material available for patellar segmentation development. Although patellar cartilage was not segmented, our early LOGISMOS approach \cite{Yin2010} was designed to segment all three bones and cartilages of the knee joint simultaneously.

\appendices

\section{Just Enough Interaction} 
\label{jei}

In this work, JEI was used to prepare data for training the proposed hierarchical RF classifier (Fig.~\ref{fig:workflow_train}) and to assess the benefits of 4D LOGISMOS over a longitudinal study period. JEI is a novel interaction technique that uses the graph based LOGISMOS framework. The user inspects the automated segmentation results and provides segmentation correction points that interact directly with the underlying graph framework. This strategy differs from traditional voxel-by-voxel editing by only requiring limited (i.e., just-enough) interactions to correct the automated LOGISMOS segmentation. The JEI interaction method still guarantees global optimality even with the cost-changes resulting from the JEI corrections.

\begin{figure}[htb]
 	{\includegraphics[width=\columnwidth]{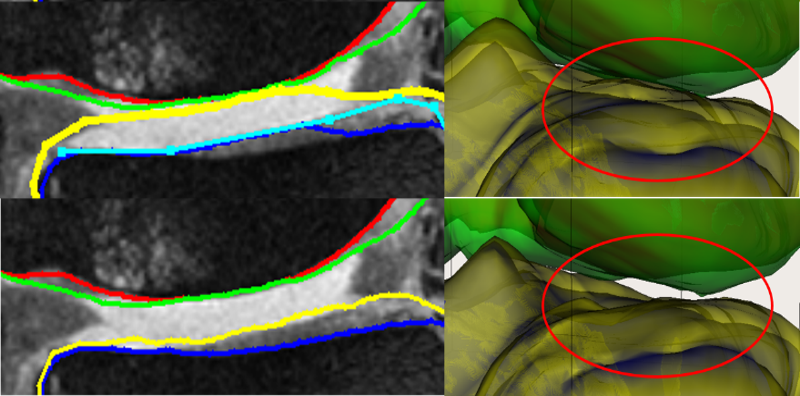}}
 	\caption{A graphical depiction of the JEI workflow showing the interactive correction steps. A single slice that is identified with improper segmentation of the tibia cartilage (yellow) is shown. The user provides the correcting set of boundary points (cyan) which identifies the 3D neighborhood of graph columns on which the costs are modified locally. The max-flow was recomputed in 3D resulting in the corrected surfaces within milliseconds. The resulting correction from JEI on a single 2D slice can be appreciated in the corresponding 3D model representation of the surface.}
 	\label{fig:jei_workflow}
\end{figure}

Following automated LOGISMOS segmentation, the JEI editing was performed on the resulting surface as needed. A custom graphical user interface (GUI) was designed onto which the image volume, residual graph and the ELF geometric graph were loaded. The JEI workflow is shown in Fig. \ref{fig:jei_workflow}. The details of the JEI architecture and GUI are given in \cite{Kashyap-IMIC2016}.
 
\section{Automated Extraction of Load-Bearing Sub-Regions}
\label{sub-plate}
Analyzing the whole cartilage structure may not be sensitive to cartilage losses that occur locally. Cartilage sub-regions have been identified as areas which bear the maximum stresses during motion with a consensus based nomenclature by the clinicians on the sub-regions of the femur and tibia that need to be analyzed separately \cite{Eckstein2006a}. Several techniques \cite{Wirth2008,Wirth2010} for identifying the sub-plates exist all of which require human interaction to provide an initialization for the sub-region analysis. We have developed a fully automated sub-plate detection algorithm that uses LOGISMOS surface meshes to define individual sub-plates.

\subsection{Trochlear Notch Identification}

\begin{figure}[htb]
		{\includegraphics*[width=0.49\columnwidth,height=4.1cm]{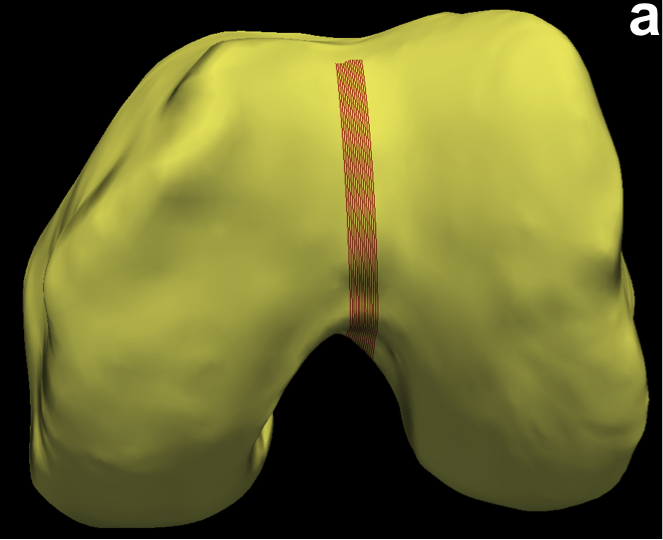}}
		{\includegraphics*[width=0.40\columnwidth]{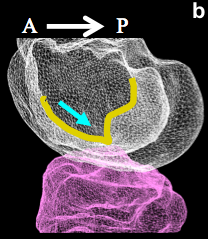}}\\	  
	\caption{Trochlear notch.  (a)~LOGISMOS-segmented bone mesh with the trochlear groove region highlighted. 
	(b)~Bone mesh highlighting the trochlear groove (overlayed in yellow) in the anterior to posterior direction. The blue arrow indicates the trochlear notch on the groove at the base of the groove curvature before the bone ridge structure rises sharply.
	}
	\label{fig:notch_eg}
\end{figure}

\begin{figure}[htb]
	\includegraphics[width=0.46\columnwidth]{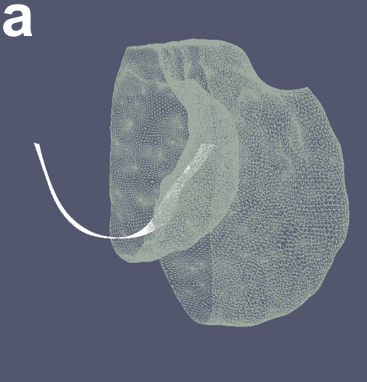}
	\includegraphics[width=0.46\columnwidth]{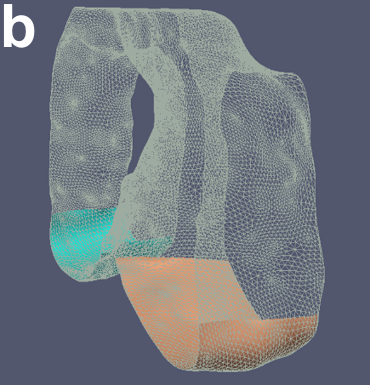}
	\caption{ Trochlear notch.  (a)~Use of AP curvature of the groove on the femoral bone. After isolating the groove region, the trochlear notch is identified by taking an average gradient on a family of closely positioned contour lines along this surface. (b)~The load-bearing regions of the femur identified by isolating 60 \% of the distance from the trochlear notch to  posterior most in the AP direction on each condyle respectively.}
	\label{fig:notch}
\end{figure}

The first step in the sub-plate extraction is the detection of the trochlear notch. The notch is at the base of a groove along which the patella (knee cap) slides over the femur (Fig.\ \ref{fig:notch_eg}). The main anatomic feature exploited is anterior to posterior (AP) curvature of the groove of the femoral bone (Fig.\ \ref{fig:notch_eg}b). The notch is at the base of the curvature before the bone ridge rises sharply. An implicit cutting plane was used to isolate the analysis region represented by $\hat{n}.p$ where $ \hat{n}$ is the normal direction from the plane and $p$ is a point on the plane.  When isolating the groove regions using simple implicit plane cutting, a family of contours were drawn along this surface. The sharp rise of the bone structure along the contour near the base of the ridge gives a large change in contour positioning value traversing in the AP direction. To increase robustness we find the positions of largest change on all of the closely positioned groove contour lines and average them to find the desired trochlear notch (Fig.\ \ref{fig:notch}). 

\subsection{Sub-Plate Detection using Implicit Cutting Plane Geometry}

Using implicit cutting planes at the trochlear notch separates the posterior region from the anterior and further splits them into the medial and lateral condyle. For each femoral condyle, the load-bearing regions is defined as 60\% of the distance in the AP direction from the trochlear notch to the posterior-most point of the respective condyles. Isolating them using cutting planes normal to the AP plane at the 60\% region isolates the load-bearing regions of the central medial and lateral femur (Fig.\ \ref{fig:notch}b).

\begin{figure}[htb]
	\centering
	\includegraphics[height=0.31\columnwidth]{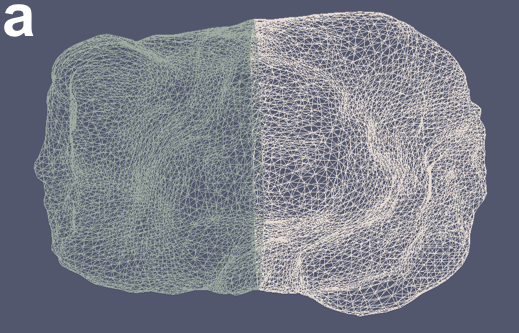}
	\includegraphics[height=0.31\columnwidth]{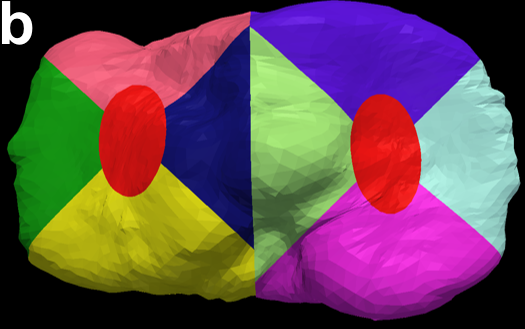}
	\caption{(a)~Isolated medial and lateral tibia, (b)~All the tibial sub-plates.}
	\label{fig:tibia}
\end{figure}

\begin{figure}[htb]
	\includegraphics[width=\columnwidth]{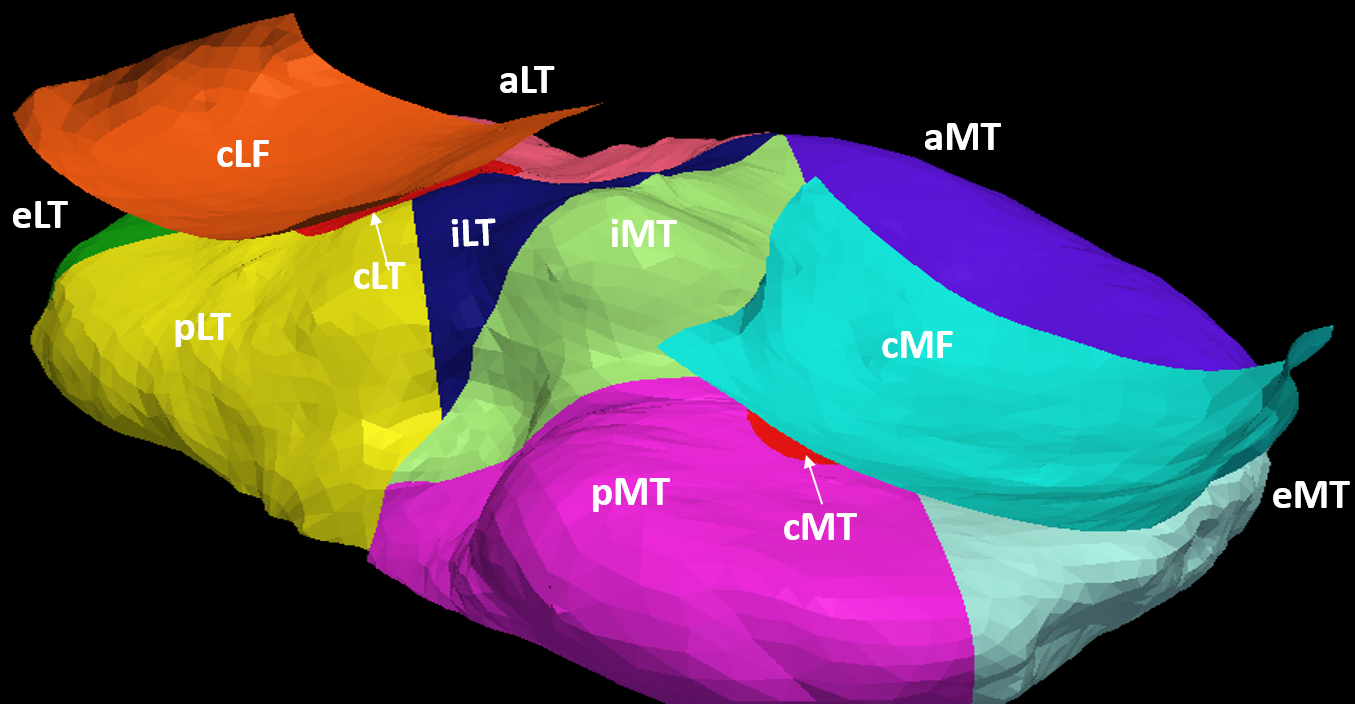}
	\includegraphics[width=\columnwidth,height=0.6\columnwidth]{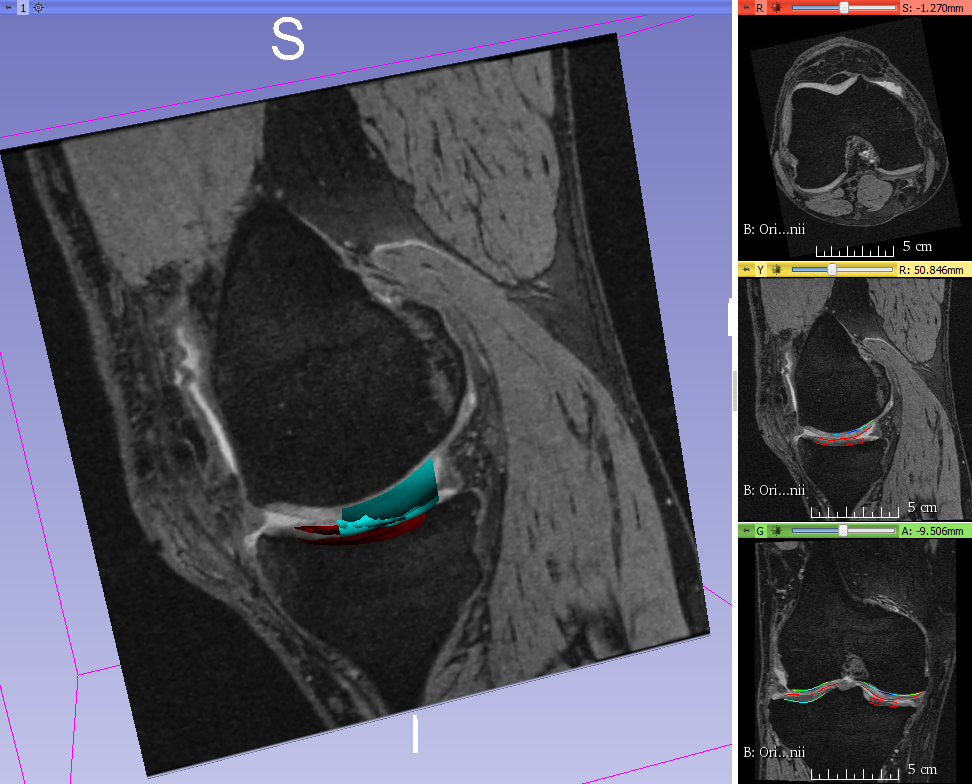}
	\caption{ Automated sub-plate division of the cartilage. The 60\% central lateral (cLF) and central medial femur (cMF) are shown. Each Medial (MT) and lateral tibia (LT) regions are subdivided as: central (cLT/cMT), interior (iLT/iMT), exterior (eLT/eMT), anterior (aLT/aMT) and posterior (pLT/pMT) regions respectively.}
	\label{fig:all_plates}
\end{figure}

The tibia is divided into the medial and lateral compartments using cutting planes positioned at the trochlear notch with its normal perpendicular to the AP plane (Fig.\ \ref{fig:tibia}). After subdividing the plates into the medial and lateral regions, we isolate the central tibia and the peripheral sub-regions. The central 20\% elliptical area of the medial and the lateral plate is computed around the center of mass of each respective plate. The radius of the major and the minor axis is computed as a ratio of bounds of the respective medial/lateral compartments. The major axis radius is computed from the compartment bounds along the AP direction and similarly, the minor axis radius is computed from the ratio of the compartment bounds perpendicular to the AP direction. Furthermore, the remaining regions are isolated using a 45$\degree$ and 135$\degree$ cutting planes around the center of mass of their respective compartments to give all four peripheral sub-regions as shown in Fig.\ \ref{fig:tibia}. The final list of all the extracted sub-plates from the automated algorithm is visually shown in Fig.\ \ref{fig:all_plates}. 

\subsection{Robustness of Sub-plate Detection}
Automated identification of sub-plate regions uses multi-surface knee joint segmentation as input. Consequently, the sensitivity of such surface segmentation on sub-plate region identification and thus on cartilage thickness measurement must be established. Load-bearing sub-plates were derived from two surface segmentation results (4D LOGISMOS, JEI-defined surfaces) and used to determine sub-plate-specific cartilage thickness from the manually segmented independent standard.
In 108 patients, the two sets of sub-plate-specific cartilage thickness measurements were compared using regression analysis, which showed high robustness (low sensitivity) of the sub-plate detection algorithm on the identification of the load-bearing regions (using manually segmented independent standard surface) with respect to the underlying knee-joint segmentation. The sub-plate-specific correlation  of the two thickness measurements ranged from $R^2 = 0.98$ to $0.99$.

\ifCLASSOPTIONcaptionsoff
  \newpage
\fi

\bibliographystyle{IEEEtranbib}
\bibliography{Final}

\end{document}